# IMLJD: A Computational Dataset for Indian Matrimonial Litigation Analysis

**Joy Bose**
Independent Researcher, Bengaluru, India

**Abstract**

We present IMLJD, an open dataset of 3,613 Indian court judgments covering matrimonial disputes under IPC Section 498A, the Protection of Women from Domestic Violence Act, and CrPC Section 482. The dataset covers the Supreme Court of India from 2000 to 2024 (1,474 cases) and the Karnataka High Court from 2018 to 2024 (2,139 cases), with structured outcome labels, metadata-derived indicators, and a knowledge graph. We find that 57.6% of quashing petitions succeed at the Supreme Court level compared to 39.7% at the Karnataka High Court level. On a matched 2018 to 2024 period, the SC quash rate is 59.3%, widening the differential to 19.6 percentage points and confirming the finding is robust to temporal adjustment. The dataset, code, and knowledge graph are released openly at https://github.com/joyboseroy/imljd and https://huggingface.co/datasets/joyboseroy/imljd.

## 1. Introduction

IPC Section 498A criminalises cruelty by a husband or his relatives against a married woman. Enacted in 1983, it addresses domestic violence and dowry harassment, both of which remain serious and underreported. At the same time, the same provision has been the subject of sustained judicial concern about mechanical arrest and vague allegations against extended family members. The Supreme Court addressed this directly in Arnesh Kumar v State of Bihar (2014), which required magistrates to apply their minds before authorising arrest, and in Rajesh Sharma v State of UP (2017), which established screening committees.

Both realities exist. Domestic violence is real and underreported. Courts have also expressed documented concern regarding procedural misuse and the mechanical implication of extended family members in matrimonial complaints. Understanding the scale and pattern of litigation requires data, not anecdote. Yet computational analysis of Indian matrimonial litigation has been limited by the lack of structured, labelled datasets. General legal NLP datasets for India such as ILDC and LawSum exist, but none focus specifically on matrimonial disputes with outcome labels and cross-court coverage.

High-volume matrimonial litigation imposes substantial procedural and emotional costs on litigants and courts alike, motivating the need for computational tools that support empirical legal analysis.

We build IMLJD to fill this gap. The dataset is built entirely from public judicial archives with a reproducible pipeline. We make no claim about whether specific allegations are true or false. The signal flags in the dataset are descriptive, not diagnostic. The framing throughout is procedural fairness research.

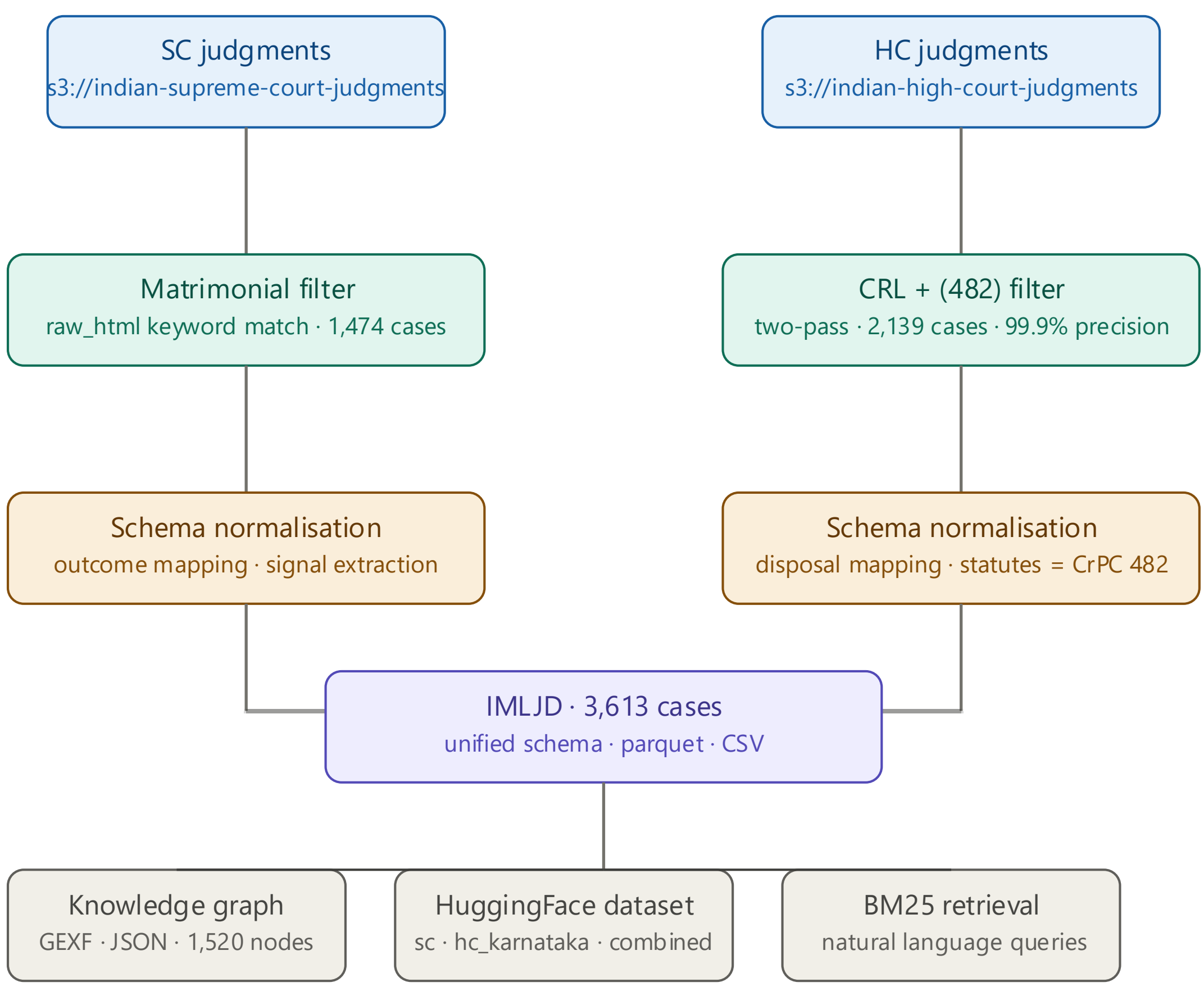

***Figure 1:*** *End-to-end IMLJD data collection and normalization pipeline. Supreme Court and Karnataka High Court judgments are independently filtered from AWS Open Data archives, normalized into a unified schema, enriched with metadata-derived indicators, and released as parquet/CSV datasets, a knowledge graph, and a BM25 retrieval interface.*

## 2. Data Sources and Collection

Both sub-corpora use publicly accessible AWS Open Data Registry datasets:

- s3://indian-supreme-court-judgments/ (parquet metadata by year, full text in tar archives)
- s3://indian-high-court-judgments/ (parquet metadata by year, court, and bench)

No credentials are required. The full collection pipeline is available in the GitHub repository.

### 2.1 Supreme Court Sub-corpus

We downloaded metadata parquet files for years 2000 to 2024 and filter using matrimonial terms applied to the raw HTML field, which contains richer text than structured metadata columns. This yields 1,474 cases. Outcomes are mapped from the disposal nature field to a standardised vocabulary: quashed, allowed, dismissed, settled, disposed, partly allowed, and unknown.

### 2.2 Karnataka High Court Sub-corpus

We apply a two-pass filter over parquet files for Karnataka HC (court ID 29 3) across three benches (Bengaluru, Dharwad, Kalaburagi):

Pass 1: CRL prefix in the title field, which identifies criminal cases. Pass 2: (482) or explicit matrimonial terms in the description field.

This yields 2,139 cases, of which 99.9% contain an explicit Section 482 reference in the judgment header, confirming near-perfect filter precision. All Karnataka HC cases are CrPC Section 482 quashing petitions.

### 3. Schema and Signal Extraction

Each record is normalised to a common schema. The SC sub-corpus includes case identifiers, party names, year, case type, outcome, statutes invoked, and a set of metadata-derived indicators. The HC sub-corpus includes case title, judge, date, disposal nature, bench, and outcome. A significant engineering contribution of IMLJD is schema normalization across heterogeneous Indian judicial archives. The two AWS Open Data sources use different parquet structures, column naming conventions, and metadata depth. The SC parquet uses a flat structure with a raw HTML field containing case summary text; the HC parquet organises files by year, court ID, and bench subdirectory with separate columns for title, description, and PDF link. Disposal nature values differ substantially across courts and years, requiring a unified outcome vocabulary (quashed, allowed, dismissed, settled, disposed, partly allowed). The normalization pipeline handles these inconsistencies automatically and is designed to extend to additional High Courts without manual schema rework.

Metadata-derived indicators are extracted by regex from available portal summary text. These should be treated as lower-bound estimates, as the underlying signals appear in full judgment text which is not available for most SC cases due to image-format PDFs. Indicators include:

- omnibus vague language: presence of "omnibus allegations", "vague allegations", or "sweeping allegations"
- relatives accused: mother-in-law, father-in-law, sister-in-law, or in-laws named as accused
- judicial criticism misuse: "misuse", "abuse of process", "legal terrorism", or "weapon of litigation"
- arnesh kumar cited / rajesh sharma cited: landmark case citations in judgment text
- mediation mentioned, settlement mentioned, arrest mentioned

A knowledge graph is built using NetworkX with node types Case, Statute, Court, Outcome, Precedent, and Year. Edge types include INVOKES, HEARD BY, RESULTS IN, CITES, and DECIDED IN. The graph contains 1,520 nodes and 13,364 edges and is released in GEXF format for Gephi and JSON for web use. The graph enables structured retrieval queries that plain text search cannot handle efficiently. For example, traversing from the Arnesh Kumar precedent node to all citing cases and filtering for quashed outcomes surfaces the subset of cases where the landmark arrest guidelines were cited and the petition succeeded. Similarly, querying all cases connected to the IPC 498A statute node and the settled outcome node identifies the 120 cases where matrimonial disputes resolved through compromise at the quash petition stage. A BM25 retrieval interface over the corpus is also released, supporting natural language queries such as "cases mentioning omnibus allegations" or "quash petitions with relatives accused" with ranked case retrieval.

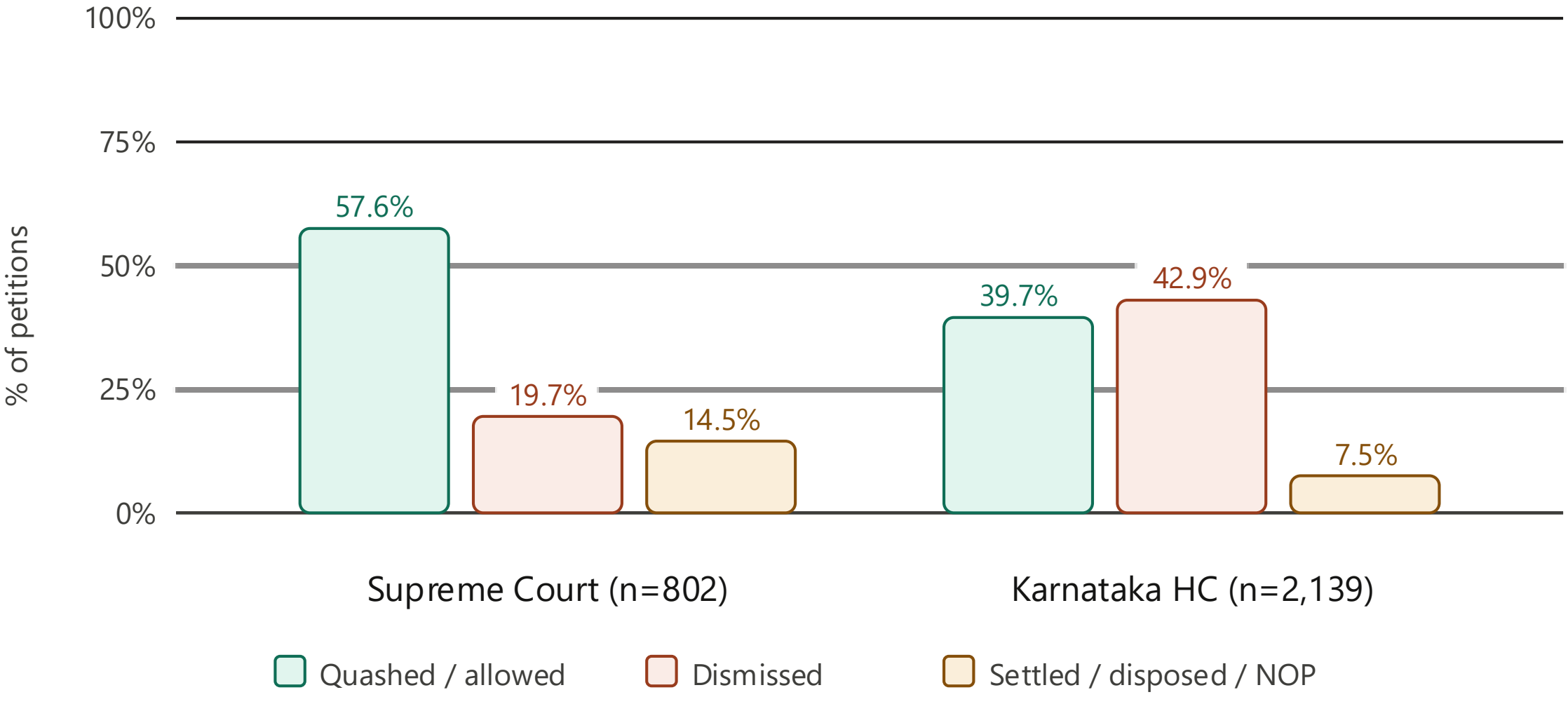

***Figure 2:*** *Distribution of quash petition outcomes across Supreme Court and Karnataka High Court sub-corpora. Supreme Court petitions show a higher quash success rate (57.6%) than Karnataka High Court petitions (39.7%), with the differential increasing to 19.6 percentage points under matched 2018–2024 temporal comparison.*

## 4. Dataset Statistics

### 4.1 Composition

Table 1 shows the composition breakdown of cases in the dataset across high court and supreme court.

**Table 1**: *Composition of the IMLJD dataset across Supreme Court and Karnataka High Court sub-corpora, including court coverage periods and total case counts.*

| **Sub-corpus** | **Cases** | **Court** | **Period** |
|---|---|---|---|
| SC matrimonial | 1,474 | Supreme Court of India | 2000 to 2024 |
| HC Karnataka | 2,139 | Karnataka High Court | 2018 to 2024 |
| Total | 3,613 | | |

### 4.2 SC Case Types

Table 2 shows the breakdown of supreme court cases within the dataset.

**Table 2:** *Distribution of Supreme Court matrimonial case types within the IMLJD corpus. Quash petitions constitute the majority category (54.4%).*

| Case Type | Count | Percent |
|---|---|---|
| Quash petitions | 802 | 54.4% |
| Appeals | 545 | 37.0% |
| Maintenance | 69 | 4.7% |
| Bail | 17 | 1.2% |
| Other | 41 | 2.8% |

**4.3 SC Outcomes**

Table 3 shows the distribution of disposal outcomes for SC cases in the dataset.

**Table 3:** *Outcome distribution for Supreme Court matrimonial litigation cases in the IMLJD dataset.*

| Outcome | Count |
|---|---|
| Quashed | 462 |
| Dismissed | 381 |
| Allowed | 247 |
| Settled or disposed | 234 |
| Partly allowed | 95 |
| Other | 55 |

**4.4 Karnataka HC Disposal**

Table 4 shows the distribution of disposal outcomes for Karnataka HC cases in the dataset.

**Table 4:** *Disposal outcomes for Karnataka High Court CrPC Section 482 matrimonial quashing petitions.*

| Disposal | Count | Percent |
|---|---|---|
| Dismissed | 917 | 42.9% |
| Allowed (quashed) | 849 | 39.7% |
| Dismissed for non-prosecution | 160 | 7.5% |

| Disposal | Count | Percent |
|---|---|---|
| Disposed (settled) | 136 | 6.4% |
| Partly allowed | 43 | 2.0% |
| Other | 31 | 1.5% |

**4.5 Statutes Invoked (SC)**

Table 5 shows the distribution of statues invoked for the SC cases in the dataset.

**Table 5:** *Frequency of explicitly indexed statutes within the Supreme Court sub-corpus metadata. Counts reflect structured metadata indexing and therefore underrepresent statutes appearing only in unstructured HTML summaries.*

| Statute | Frequency |
|---|---|
| IPC 498A | 194 |
| Dowry Prohibition Act | 60 |
| Hindu Marriage Act | 48 |
| IPC 304B | 44 |
| CrPC 482 | 41 |
| DV Act | 23 |
| CrPC 125 | 9 |

**5. Key Findings**

**5.1 Quash Success Rate Differential**

57.6% of quashing petitions succeed at the Supreme Court level (462 of 802), compared to 39.7% at the Karnataka High Court level (849 of 2,139). To account for the time period difference between the two sub-corpora, we also compute the SC quash rate for 2018 to 2024 only, matching the HC coverage period. The SC quash rate for 2018 to 2024 is 59.3% (146 of 246), widening the differential to 19.6 percentage points. The finding is therefore robust to temporal adjustment. This differential is consistent with appellate pre-filtering: cases that reach the SC on quashing have already been considered by a High Court, selecting for stronger arguments. It may also reflect differences in judicial approach across court levels, and the significant resources required to pursue litigation to the apex court. Disentangling these explanations requires a matched longitudinal study, which the IMLJD pipeline supports.

**5.2 Settlement at Quash Stage**

122 of 802 SC quash petitions (15.2%) resolve through settlement or compromise rather than adjudication. This is a notable proportion and suggests that the filing of a quash petition itself creates conditions for settlement, independent of the judicial outcome.

### 5.3 Year Distribution

SC caseload peaks in 2007 to 2009 (around 110 cases per year) and declines gradually afterward. This is consistent with increased 498A filings in the mid-2000s and subsequent judicial interventions reducing litigation volume.

## 6. Limitations

The limitations of the work include the following:

**Metadata-level filtering**: The SC matrimonial filter uses the raw HTML field from the SC portal, which contains case summaries rather than full judgment text. Some non-matrimonial cases may have been included. A statute-confirmed subset of 322 SC cases with explicit statute references in metadata shows a similar quash success rate (57.9%), suggesting the headline finding is robust. The statute frequency counts in Section 4.5 reflect cases where the statute was explicitly indexed in structured metadata fields. The remaining cases were captured via text-mining of HTML summaries where statutes were mentioned but not formally indexed, which accounts for the lower explicit IPC 498A count relative to total cases.

**Single HC jurisdiction**: The HC sub-corpus covers only Karnataka. Different High Courts may have different filing patterns and judicial approaches.

**Image-format PDFs**: SC judgments in the tar archives are scanned image PDFs and cannot be parsed for full text without OCR. Signal flags for the SC sub-corpus are therefore extracted from the metadata HTML field, which is shorter and less complete than the full judgment text.

**Ground truth ambiguity**: A quashed FIR does not mean the underlying allegation was false. A dismissed petition does not mean it was well-founded. All findings should be interpreted procedurally, not substantively.

## 7. Related Work

ILDC (Malik et al., 2021) provides approximately 35,000 Supreme Court cases for outcome prediction and explainability research. LawSum (Shukla et al., 2022) provides summaries for 10,000 judgments. InLegalNLP provides benchmark NLP tasks for Indian legal text. IMLJD complements these by focusing on matrimonial litigation with domain-specific outcome labels, signal flags, and a knowledge graph.

This work is part of a broader research direction on graph-grounded legal reasoning, described in Bose (2026) with the FalkorDB-IRAC system.

## 8. Conclusion and Future work

IMLJD provides the first structured, labelled, and reproducible computational dataset for Indian matrimonial litigation, built from public judicial archives. The dataset surfaces a 19.6-point quash success rate differential between SC and HC levels on a matched time period, a 15% settlement rate at the quash petition stage, and year-level trends in litigation volume. All data, code, and the knowledge graph are released openly. The pipeline is designed for extension to additional High Courts and future years. Researchers using this dataset should note that quash success rates do not indicate the veracity of underlying allegations. A quashed FIR reflects a procedural determination, not a factual finding. The metadata-derived indicators are descriptive starting points for hypothesis generation, not validated classifiers. Downstream applications that use this dataset to argue that matrimonial complaints are systematically false would misrepresent both the data and the legal process.

Future work includes OCR-based extraction of full-text Supreme Court judgments, extension to additional High Courts, rhetorical-role labelling, and citation-outcome analysis. Full-text extraction would enable deeper analysis of judicial reasoning patterns, including procedural grounds for dismissal, maintenance rejection, and settlement-linked quashing. Integration with APIs such as Indian Kanoon or improved OCR pipelines may support richer precedent analysis and temporal litigation pathways.

**References**

1. Arnesh Kumar v State of Bihar, (2014) 8 SCC 273.
2. Rajesh Sharma v State of UP, (2017) 15 SCC 133.
3. Malik et al. ILDC for NLP: Indian Legal Documents Corpus for Court Judgment Prediction. ACL 2021.
4. Shukla et al. LawSum: A Weakly Supervised Approach for Indian Legal Document Summarization. 2022.
5. Bose, J. FalkorDB-IRAC: Graph-Constrained Generation for Verified Legal Reasoning in Indian Judicial AI. arXiv:2605.14665, 2026.
6. AWS Open Data Registry. Indian High Court Judgments. https://registry.opendata.aws/indian-high-court-judgments/
7. AWS Open Data Registry. Indian Supreme Court Judgments. https://registry.opendata.aws/indian-supreme-court-judgments/

**Data and Code Availability**

Dataset: https://huggingface.co/datasets/joyboseroy/imljd

Code: https://github.com/joyboseroy/imljd